\documentclass{article}

\PassOptionsToPackage{square,numbers,sort&compress}{natbib}

\usepackage[preprint]{neurips_2023}

\usepackage[utf8]{inputenc} 
\usepackage[T1]{fontenc}    
\usepackage{hyperref}       
\usepackage{url}            
\usepackage{booktabs}       
\usepackage{amsfonts}       
\usepackage{nicefrac}       
\usepackage{microtype}      
\usepackage{xcolor}         

\usepackage{amsmath}
\usepackage{amssymb}
\usepackage{amsthm}
\newtheorem{theorem}{Theorem}
\newtheorem{remark}{Remark}
\usepackage{algorithm}
\usepackage{algorithmic}
\usepackage{graphicx}
\graphicspath{{./output/}}
\DeclareGraphicsExtensions{.png}
\def\Appref#1{supplementary materials}

\usepackage{amsmath,amsfonts,bm}

\def\eqref#1{equation~\ref{#1}}

\def\1{\bm{1}}

\def\vzero{{\bm{0}}}

\def\vtheta{{\bm{\theta}}}

\def\vb{{\bm{b}}}

\def\vt{{\bm{t}}}

\def\vx{{\bm{x}}}

\def\vz{{\bm{z}}}

\def\mA{{\bm{A}}}

\def\mI{{\bm{I}}}

\def\mM{{\bm{M}}}

\def\mP{{\bm{P}}}

\def\mW{{\bm{W}}}
\def\mX{{\bm{X}}}
\def\mY{{\bm{Y}}}
\def\mZ{{\bm{Z}}}

\DeclareMathAlphabet{\mathsfit}{\encodingdefault}{\sfdefault}{m}{sl}
\SetMathAlphabet{\mathsfit}{bold}{\encodingdefault}{\sfdefault}{bx}{n}

\newcommand{\Ls}{\mathcal{L}}

\title{Unifying Label-inputted Graph Neural Networks with Deep Equilibrium Models}

\author{%
  Yi Luo\\
  University of Electronic Science and Technology of China\\
  \texttt{cf020031308@163.com}
  \And
  Guiduo Duan\\
  University of Electronic Science and Technology of China\\
  \And
  Guangchun Luo\\
  University of Electronic Science and Technology of China\\
  \And
  Aiguo Chen\thanks{Correspongin Author: Aiguo Chen}\\
  University of Electronic Science and Technology of China\\
  \texttt{agchen@uestc.edu.cn}
}

\begin{document}

\maketitle

\begin{abstract}
    The success of Graph Neural Networks (GNN) in learning on non-Euclidean data arouses many subtopics, such as Label-inputted GNN (LGNN) and Implicit GNN (IGNN).
    LGNN, explicitly inputting supervising information (a.k.a.\ labels) in GNN, integrates label propagation to achieve superior performance, but with the dilemma between its propagating distance and adaptiveness.
    IGNN, outputting an equilibrium point by iterating its network infinite times, exploits information in the entire graph to capture long-range dependencies, but with its network constrained to guarantee the existence of the equilibrium.
    This work unifies the two subdomains by interpreting LGNN in the theory of IGNN and reducing prevailing LGNNs to the form of IGNN\@.
    The unification facilitates the exchange between the two subdomains and inspires more studies.
    Specifically, implicit differentiation of IGNN is introduced to LGNN to differentiate its infinite-range label propagation with constant memory, making the propagation both distant and adaptive.
    Besides, the masked label strategy of LGNN is proven able to guarantee the well-posedness of IGNN in a network-agnostic manner, granting its network more complex and thus more expressive.
    Combining the advantages of LGNN and IGNN, Label-inputted Implicit GNN (LI-GNN) is proposed.
    It can be widely applied to any specific GNN to boost its performance.
    Node classification experiments on two synthesized and six real-world datasets demonstrate its effectiveness.
    Code is available at \url{https://github.com/cf020031308/LI-GNN}.
\end{abstract}

\section{Introduction}

Node classification is a task widely occurring in real-world scenarios with graphic data, such as traffic speed forecasting~\cite{DBLP:conf/ijcai/WuPLJZ19}, field recommendation for article submissions~\cite{DBLP:conf/nips/HuFZDRLCL20}, image classification~\cite{Liu2019PrototypePN}, and molecular property prediction~\cite{Gilmer2017NeuralMP}.
The goal is to assign predefined labels to graph nodes according to node features and the graph structure.
In recent years, Graph Neural Network (GNN)~\cite{DBLP:journals/tnn/WuPCLZY21} has become one of the most dominant methods for classifying graph nodes.
It relies on deep learning techniques to capture hidden data patterns in an end-to-end scheme, where node features and the graph structure are at the input end while known labels supervise the output end.
The success of GNN derives many different lines of work focusing on varying improvements, such as Label-inputted GNN (LGNN) and Implicit GNN (IGNN)~\cite{DBLP:conf/nips/GuC0SG20}.

LGNN improves the classification performance of GNN by explicitly inputting known labels~\cite{DBLP:journals/corr/abs-2002-06755} and utilizing their structural information.
It combines the advantages of label propagation and learnable feature propagation, producing state-of-the-art performance on various benchmarks.
However, it is unaffordable to grant long-range propagation with deep learning ability due to expensive memory consumption, leaving the balance between the distance of label propagation and its adaptiveness an open question.
IGNN improves the long-range dependencies capturing ability of its backbone GNN (also known as implicit layer) by repeatedly inputting its output until the output converges to gather distant information.
It utilizes implicit differentiation to optimize the converged equilibrium, consuming only constant memory without taping the gradients during intermediate loops.
However, the backbone GNN have to be well-posed to guarantee the existence and uniqueness of the equilibrium point, resulting in its simple structure and weak expressiveness.

This work represents LGNN in IGNN's form by treating pseudo labels of the former as the targetted equilibrium of the latter.
The unification can facilitate the exchange between LGNN and IGNN.
Any specific LGNN with parametric propagation can integrating implicit differentiation of IGNN to derive a new method, which is both distant and adaptive in propagating labels.
Besides, the masked label strategy (MLS) of LGNN is proven to guarantee IGNN's well-posedness with a proper mask rate.
Since MLS is network-agnostic, its granted IGNN can have complex backbone GNN, such as any state-of-the-art model, to gain expressiveness without worrying about its well-posedness condition.
Combining the advantages of LGNN and IGNN, Label-inputted Implicit GNN (LI-GNN) is proposed.
Node classification experiments on two synthesized datasets demonstrated that LI-GNN inherits label utilization ability from LGNN and long-range propagation ability from IGNN\@.
Experiments on real-world datasets further verify its effectiveness.

\section{Related works}

This section explores typical methods of GNN and its two subdomains, including Label-inputted GNN and Implicit GNN.

\subsection{Notations}

In the discussion of this article, bias terms in neural networks are omitted for clarity because a formula with a bias term as $\vx \mW + \vb$ is equivalent to $\begin{bmatrix} \vx & 1 \end{bmatrix} \begin{bmatrix} \mW & \vb \end{bmatrix}^T$ without a bias term.
Matrices appearing in partial derivatives stand for their vectorized forms by chaining their rows, such as the $\mZ$ in $\frac{\partial f_\vtheta(\mZ)}{\partial \mZ}$.

\subsection{Graph Neural Network}

A Graph Neural Network (GNN) method is based on end-to-end deep learning techniques.
It inputs graph information, including its adjacency matrix $\mA$ and node feature matrix $\mX$, into GNN and supervises its output to capture data patterns.
Supervising information in node classification tasks is a given 0-1 matrix $\mY$.

Graph Convolutional Network (GCN)~\cite{DBLP:journals/corr/KipfW16} is one of the most commonly used GNNs.
It recursively convolutes node information along the graph structure to learn node representations.
Denoting an $L$-layered GCN as $\mbox{GCN}^{(L)}(\mA, \mX)$, its output $\mZ^{(L)}$ is recursively calculated as
\[
    \mZ^{(i+1)} = \sigma^{(i)}(\tilde \mA \mZ^{(i)} \mW^{(i)}), i = 0, 1, \ldots, L - 1,
    \quad
    \mZ^{(0)} = \mX,
\]
where $\tilde \mA$ is a normalized matrix of $\mA$, $\mW^{(i)}$ is learnable, and $\sigma^{(i)}$ is a nonlinear activator.

Simple Graph Convolution (SGC)~\cite{DBLP:conf/icml/WuSZFYW19} simplifies all but the last activators $\sigma^{(i)}, i = 0, 1, \ldots, L - 2$ of GCN to be linear and represents the model as $\mZ^{(i+1)} = \sigma(\tilde \mA^L \mX \mW)$.
$\tilde \mA^L \mX$ is precomputed and reused in training, making SGC significantly efficient.

\subsection{Label-inputted GNN}

A Label-inputted GNN (LGNN) method inputs known ground-truth labels $\mM_t \mY$ into its backbone GNN to predict unknown labels $\mM_{\bar t} \mY$, as
\begin{equation}\label{eq:lgnn0}
    \hat \mY = f_\vtheta(\mM_t \mY, \mA, \mX),
\end{equation}
where $f_\vtheta$ is the backbone GNN, $\mM_t$ is a diagonal 0-1 matrix indicating nodes in the training set, and $\mM_{\bar t} = \mI - \mM_t$ indicates nodes beyond the training set.
In some LGNN methods, the predictions $\hat \mY$ overwritten by ground-truth labels are recursively inputted into the GNN, making
\begin{equation}\label{eq:lgnn}
    \hat \mY = f_\vtheta(\tilde \mY_t, \mA, \mX),
    \quad
    \tilde \mY_t = \mM_{\bar t} \hat \mY + \mM_t \mY,
\end{equation}
where $\tilde \mY_t$ is termed as pseudo labels with respect to the training set indicated by $\mM_t$.

Specifically, Label Propagation Algorithm (LPA)~\cite{DBLP:conf/icml/ZhuGL03} propagates known labels along edges repeated till convergence, as
\begin{equation}\label{eq:lpa}
    \hat \mY = \tilde \mA_\lambda \tilde \mY_t,
    \quad
    \tilde \mA_\lambda = (1 - \lambda) \mI + \lambda \tilde \mA,
\end{equation}
where $\lambda$ is a hyperparameter.
ResLPA~\cite{DBLP:conf/iccai/LuoHCZ21} approximates residuals between labels of adjacent nodes and corrects pseudo labels $\tilde \mY_t$.
3ference~\cite{Luo2022InferringFR} adaptively learns label transition patterns by inputting them into its backbone $\mbox{GCN}^{(1)}$ to avoid unreasonable heuristic assumptions.
Correct and Smooth (C\&S)~\cite{DBLP:conf/iclr/HuangHSLB21} runs LPA~(\ref{eq:lpa}) two times after a base prediction $\hat \mY_0$ is made by a deep learning model like GNN or Multi-Layered Perceptron (MLP).
The first LPA with hyperparameter $\lambda_c$ in the Correct phase is to propagate residuals between known labels and the base predictions from the training set $\mM_t (\mY - \hat \mY_0)$ to the entire graph.
The second LPA with hyperparameter $\lambda_s$ in the Smooth phase is to propagate corrected predictions for smoothing purposes.
Label Input~\cite{DBLP:conf/ijcai/ShiHFZWS21} leverages masked label strategy to randomly split the known labels $\mM_t \mY$ into two exclusive parts in every training epoch, inputting one part $\mM_i \mY$ into GNN and leaving another part $\mM_o \mY$ as supervising information, where $\mM_o = \mI_o \mM_t$, $\mM_i = \mM_t - \mM_o$, $\mI_o$ is a diagonal 0-1 matrices satisfying $\| \mI_o \|_1 \approx \alpha \| \mI \|_1 $, and $\alpha < 1$ is the mask rate.
Label Reuse~\cite{DBLP:journals/corr/abs-2103-13355} is stacked upon Label Input.
It iterates the backbone GNN multiple times starting from masked labels $\mM_i \mY$ to produce pseudo labels $\tilde \mY_i = \mM_i \mY + (\mM_{\bar \vt} + \mM_o) \hat \mY$ which are more informative.

A constraint of existing LGNN methods is that the label propagation is either pattern-fixed or distance-limited.
In LPA, ResLPA, or any phase of C\&S, the propagation is defined by the hyperparameter $\lambda$ which is manually tuned and thus inadaptive to varying datasets.
In 3ference and Label Input, label propagations are optimizable but their distances are restricted by numbers of GNN layers, which are again limited due to the over-smoothing problem~\cite{DBLP:conf/iccv/Li0TG19} and expensive optimizing consumptions.
In Label Reuse, pseudo labels are iteratively updated in the forward pass and the propagation distance is theoretically unlimited.
However, due to the increasing memory footprint of iterations, only the last iteration is optimized in the backward pass.
Such asymmetric between the forward pass and the backward pass is inadequate to guarantee that the convergence is a locally optimal solution.

\subsection{Implicit GNN}

An Implicit GNN (IGNN)~\cite{DBLP:conf/nips/GuC0SG20} method learns node representations $\mZ$ by solving an equilibrium equation with respect to $\mZ$, as
\begin{equation}\label{eq:ignn}
    \mZ = f_\vtheta(\mZ, \mA, \mX),
\end{equation}
where $f_\vtheta$ is a GNN.
In this work, $f_\vtheta(\mZ, \mA, \mX)$ is also abbreviated as $f_\vtheta(\mZ)$ for clarity.
To avoid confusion, IGNN denotes the abstract model defined by Equation~\ref{eq:ignn} and IGNN* denotes the specific model proposed in~\citet{DBLP:conf/nips/GuC0SG20} with
\begin{equation}\label{eq:ignn0}
    f_\vtheta(\mZ, \mA, \mX) = \sigma(\tilde \mA \mZ \mW_z + \mX \mW_x).
\end{equation}

In the forward pass, the solution to Equation~\ref{eq:ignn} is obtained by an arbitrary black-box root finder such as Broyden's method~\cite{broyden1965class} or naive forward iteration.
The latter is to iterate Equation~\ref{eq:ignn} until the convergence of $\mZ$ if the convergence is guaranteed.
Then, the backward pass is accomplished with implicit differentiation~\cite{DBLP:conf/nips/BaiKK19}.
In detail, the gradient $\nabla_\mZ \Ls$ accumulated at $\mZ$ back-propagated from the scalar loss $\Ls(\mZ)$ is multiplied by the inverse Jacobian of $\mZ - f_\vtheta(\mZ)$ before propagating to network weights $\vtheta$, as
\[
    \nabla_\mZ \Ls = {(\mI - \frac{\partial{f_\vtheta(\mZ)}}{\partial{\mZ}})}^{-T} \cdot \frac{\partial{\Ls(\mZ)}}{\partial{\mZ}},
    \quad
    \nabla_\vtheta \Ls = \frac{\partial{f_\vtheta(\mZ)}}{\partial{\vtheta}} \cdot \nabla_\mZ \Ls.
\]
Since only the last step of naive forward iteration $\Ls(f_\vtheta(\mZ))$ is taped by the automatic differentiation system, the gradient descent consumes only constant memory regardless of how many steps it takes to reach the equilibrium $\mZ$ in the forward pass.

To avoid the expensive calculation of the inverse Jacobian, IGNN* proposes to obtain the gradient $\nabla_{\mZ}$ by solving another equilibrium equation, as
\begin{equation}\label{eq:id}
    \nabla_\mZ \Ls = \frac{\partial{f_\vtheta(\mZ)}}{\partial{\mZ}} \cdot \nabla_\mZ \Ls + \frac{\partial{\Ls(\mZ)}}{\partial{\mZ}}.
\end{equation}
A recent study named Efficient IGNN (EIGNN)~\cite{DBLP:conf/nips/LiuKHWX21} derives a closed-form solution to the inverse Jacobian with matrix decomposition, avoiding numeric loss and non-convergence issues in iterative solvers.

A crucial requisite of IGNN methods is the well-posedness condition which guarantees that the forward iteration can eventually converge to an equilibrium~\cite{DBLP:conf/nips/WinstonK20}.
This severely restricts the flexibility of designing the backbone GNN in IGNN.
Existing IGNN methods only adopt simple-structured networks to be their backbones, making them difficult to address complex tasks.
For example, the backbone of IGNN* is simpler than a $\mbox{GCN}^{(1)}$ and that of EIGNN is furtherly simplified.

\section{Methodology}

In this section, LGNN and its six prevailing implementations are rewritten into the form of IGNN\@.
Their bridging benefits each other.
Implicit differentiation of IGNN is applied to LGNN with parametric propagation to lengthen its distance.
The masked label strategy of LGNN is applied to IGNN with arbitrary backbone GNN to guarantee its well-posedness. 
Then, Label-inputted Implicit GNN (LI-GNN) is proposed to combine the advantages of both LGNN and IGNN\@.

\subsection{Interpreting LGNN with IGNN}

This subsection demonstrates that LGNN is a special form of IGNN by rewriting LGNN~(\ref{eq:lgnn0},~\ref{eq:lgnn}) as
\begin{equation}\label{eq:lignn}
    \tilde \mY_t = \mM_{\bar t} f_\vtheta(\mM \tilde \mY_t, \mA, \mX) + \mM_t \mY,
\end{equation}
where $\mM = \mM_t \mbox{ or } \mI$.
The right part of Equation~\ref{eq:lignn} is an implementation of the backbone network in IGNN~(\ref{eq:ignn}) with its equilibrium $\mZ$ equals to pseudo labels $\tilde \mY_t$.
Thus, an LGNN method is also an IGNN\@.

\begin{table}\small
\caption{Rewriting LGNN methods in the form of IGNN}\label{tbl:lgnns}
\centering
\begin{tabular}{ll}
\toprule
    Method & LGNN and its rewritten IGNN \\
\midrule
    LPA & $\hat \mY = \tilde \mA_\lambda (\mM_{\bar t} \hat \mY + \mM_t \mY)$ \\
    & $\tilde \mY_t = \mM_{\bar t} \tilde \mA_\lambda \tilde \mY_t + \mM_t \mY$ \\
\midrule
    ResLPA & $\hat \mY = \tilde \mA_\lambda (\mM_{\bar t} \hat \mY + \mM_t \mY) + \lambda f_\vtheta(\mA, \mX)$ \\
    & $\tilde \mY_t = \mM_{\bar t} \tilde \mA_\lambda \tilde \mY_t + \lambda \mM_{\bar t} f_\vtheta(\mA, \mX) + \mM_t \mY$ \\
\midrule
    3ference & $\hat \mY = f_\vtheta(\tilde \mA - \mbox{diag}(\tilde \mA), \begin{bmatrix} \mM_{\bar t} \hat \mY + \mM_t \mY & \mX \end{bmatrix}) $ \\
        & $\tilde \mY_t = \mM_{\bar t} f_\vtheta(\mA - \mbox{diag}(\mA), \begin{bmatrix} \tilde \mY_t & \mX \end{bmatrix}) + \mM_t \mY$ \\
\midrule
    Correct of C\&S & $\hat \mY = \tilde \mA_{\lambda_c} (\mM_{\bar t} (\hat \mY - \hat \mY_0) + \mM_t (\mY - \hat \mY_0)) + \hat \mY_0$ \\
    & $\tilde \mY_t = \mM_{\bar t} \tilde \mA_{\lambda_c} \tilde \mY_t + \mM_{\bar t} (\mI - \tilde \mA_{\lambda_c}) \hat \mY_0 + \mM_t \mY$ \\
\midrule
    Label Input (SGC) & $\hat \mY = \mbox{SGC}^{(L)}(\mA, \begin{bmatrix} \mM_i \mY & \mX \end{bmatrix})$ \\
        & $\tilde \mY_t = \mM_{\bar t} \sigma([{\tilde \mA}^L - \mbox{diag}({\tilde \mA}^L)] \mM_t \tilde \mY_t \mW_y + {\tilde \mA}^L \mX \mW_x) + \mM_t \mY$ \\
\midrule
    Label Input (GNN) & $\hat \mY = f_\vtheta(\mA, \begin{bmatrix} \mM_i \mY & \mX \end{bmatrix})$ \\
        & $\tilde \mY_i = (\mM_{\bar t} + \mM_o) f_\vtheta(\mA, \begin{bmatrix} \mM_i \tilde \mY_i & \mX \end{bmatrix}) + \mM_i \mY$ \\
\midrule
    Label Reuse (GNN) & $\hat \mY = f_\vtheta(\mA, \begin{bmatrix} (\mM_{\bar t} + \mM_o) \hat \mY + \mM_i \mY & \mX \end{bmatrix})$ \\
        & $\tilde \mY_i = (\mM_{\bar t} + \mM_o) f_\vtheta(\mA, \begin{bmatrix} \tilde \mY_i & \mX \end{bmatrix}) + \mM_i \mY$ \\
\bottomrule
\end{tabular}
\end{table}

Specifically, existing LGNN methods, including LPA, ResLPA, 3ference, the Correct phase of C\&S, Label Input, and Label Reuse can be rewritten in the form of IGNN~(\ref{eq:lignn}) in Table~\ref{tbl:lgnns}.
Details of the conversions are in \Appref{app:lgnns}.
As can be seen from the table, LPA, ResLPA, and the Correct phase of C\&S also have the same form as IGNN*~(\ref{eq:ignn0}), if omitting the bias term $\mM_t \mY$, treating $\mM_{\bar t} \tilde \mA_\lambda$ or $[\tilde \mA - \mbox{diag}(\tilde \mA)] \mM_t$ in a whole as a normalized adjacency matrix, and regarding the rest part as node representations that are outputted by an upstream network.
Moreover, if the activator $\sigma(\cdot)$ is element-wise, $\mM_{\bar t}$ can pass through the activator to convert Label Input for SGC to the form of IGNN*, too.
For example, if $\sigma(\cdot)$ is LeakyReLU, $\mM_{\bar t}$ can be moved from the outside of $\sigma(\cdot)$ to its inside.
If $\sigma(\cdot)$ is Sigmoid, $\mM_{\bar t}$ can move in with zeros in its diagonal line replaced by $-\infty$.

\subsection{Long-range adaptive LGNN with implicit differentiation}

Unifying LGNN and IGNN can help to resolve issues in one subdomain by importing solutions from another.
In LGNN, label propagations are limited by either pattern or distance.
On the contrary, propagations in IGNN are both adaptive and infinitely distant thanks to implicit differentiation.
Since LGNN is reduced to IGNN, it is straightforward to introduce implicit differentiation~(\ref{eq:id}) to LGNN methods with learnable propagations to enchant the propagation range infinitely long.
Take Label Reuse as an example, in its forward pass, predictions are iteratively updated by the backbone GNN and overwritten by a part of known labels.
Gradients of the last iteration are taped by the automatic differentiation system.
In its backward pass, instead of using vanilla differentiation $\nabla_{\mZ} = \frac{\partial{\mathcal{L}(\mZ)}}{\partial{\mZ}}$ as in the original method, implicit differentiation~(\ref{eq:id}) can be adopted to get the equilibrium gradients.
In such a way, propagations in the forward pass and the backward pass are both infinitely long-range.
Such symmetric avoid the bias in minimizing the loss of Label Reuse.

\subsection{Well-posed IGNN with masked label strategy}

Another insight from Table~\ref{tbl:lgnns} is the difference between Label Input and Label Reuse in terms of label leakage.
While the inputted state $\mM_i \tilde \mY_i = \mM_i \mY$ of Label Input is constant and contains no information about the supervising labels $\mM_o \mY$, the inputted state $\tilde \mY_i = \mM_{\bar t} \hat \mY + \mM_o \hat \mY + \mM_i \mY$ of Label Reuse contains predictions $\mM_o \hat \mY$ made on the supervising nodes.
As the training progresses, the predictions $\mM_o \hat \mY$ become increasingly close to the supervising labels $\mM_o \mY$.
Feeding them during training causes severe information leakage, as the backbone GNN $f_\vtheta$ can degenerate to meet $\frac{\partial f_\vtheta(\mZ)}{\partial \mZ} \approx \mI$ and simply output inputted state to avoid increasing the loss.
This label leakage issue may occur in any LGNN method defined by Equation~\ref{eq:lignn}, resulting in predictions without generalization and ultimately leading to performance degradations.
From the aspect of IGNN, label leakage violates the well-posedness because of infinitely many equilibriums.
In turn, guaranteeing well-posedness can prevent performance loss brought by leaked labels because $\frac{\partial f_\vtheta(\mZ)}{\partial \mZ}$ is pushed away from $\mI$ to keep the Jacobian of $\mZ - f_\vtheta(\mZ)$ with respect to $\mZ$ invertible.
However, existing theoretical guarantees of well-posedness are only applicable to IGNN with simple-structured backbones such as IGNN*.

To guarantee well-posedness without restricting backbone networks, this work demonstrates that masked label strategy (MLS), a technique to prevent label leakage in LGNN, can also guarantee well-posedness once applied to the looping state.
According to theorems in \citet{DBLP:journals/corr/abs-2110-07190}, the implicit goal of label-masked IGNN with a $\mbox{GCN}^{(1)}$ backbone is to optimize the backbone with a regularization term that is proportional to $\| \mW_z \|_F^2$, and a large hyperparameter of the label mask rate $\alpha = \| \mM_o \|_1 / \| \mM_t \|_1$ can suppress $\| \mW_z \|_F^2$ during training.
Meanwhile, small $\| \mW_z \|_F^2$ in IGNN with a $\mbox{GCN}^{(1)}$ backbone guarantees its well-posedness, as the following theorem goes.
\begin{theorem}\label{thm:wellposed}
    With $\| \mW_z \|_F^2 < 1 / n $, where $n$ is the number of nodes, IGNN with a one-hop convolutional backbone is well-posed.
\end{theorem}
\noindent
The proof is in \Appref{app:wellposed}.
Combining the aforementioned information, if MLS is applied to the looping state and the ratio $\alpha$ of masked labels is large enough, iterating IGNN with a $\mbox{GCN}^{(1)}$ backbone from an arbitrary start point is theoretically guaranteed to converge to an existing and unique equilibrium.

This conclusion still holds when the backbone is arbitrarily complex.
As the ratio grows, the number of inputted labels progressively approaches near zero.
Then fewer labels are inputted to make a prediction.
Consequently, the inputted labels have almost no impact on the output, thereby yielding a unique and stable equilibrium for any start point.
To the extreme, if the mask rate is 100\%, both the forward pass and the backward pass of LGNN are well-posed with the equilibriums as $f_\vtheta(\mA, \begin{bmatrix} 0 & \mX \end{bmatrix})$ and $\frac{\partial{\Ls (\mZ)}}{\partial \mZ}$, respectively.
Thus, MLS with a large ratio of labels masked also guarantees the well-posedness of IGNN with an arbitrary backbone.

\subsection{Label-inputted implicit GNN (LI-GNN)}

\begin{algorithm}[!t]
    \caption{Label-inputted implicit GNN (LI-GNN)}\label{algo:lignn}
    \begin{algorithmic}[1]
        \REQUIRE GNN $f_\vtheta$, adjacency matrix $\mA$, feature matrix $\mX$, training set indicator $\mM_t$, ground-truth labels $\mY$, forward mask flag, label mask rates $\beta$, learning rate $\eta$.
        \STATE Randomly split $\mI = \mI_i + \mI_o$ to make $\| \mI_o \|_1 \approx \beta \| \mI \|_1$.

        \STATE $\hat \mY := \vzero$
        \REPEAT[Forward pass to obtain the equilibrium state]
            \IF{forward mask flag}
                \STATE $\hat \mY = f_\vtheta(\mA, \mI_i (\begin{bmatrix} (\mI - \mM_t) \hat \mY + \mM_t \mY) / (1 - \beta) & \mX \end{bmatrix})$
            \ELSE
                \STATE $\hat \mY = f_\vtheta(\mA, \begin{bmatrix} (\mI - \mM_t) \hat \mY + \mM_t \mY & \mX \end{bmatrix})$
            \ENDIF
        \UNTIL{$\hat \mY$ converges or iterations exhaust}

        \STATE $\tilde \mY_i := \mI_i ((\mI - \mM_t) \hat \mY + \mM_t \mY) / (1 - \beta)$ \COMMENT{Masked label strategy}

        \STATE $\tilde \mY := f_\vtheta(\mA, \begin{bmatrix} \tilde \mY_i & \mX \end{bmatrix})$
        \STATE $\nabla_\mZ \Ls := 0$
        \REPEAT[Backward pass with implicit differentiation]
            \STATE $\nabla_\mZ \Ls = \frac{\partial{\tilde \mY}}{\partial{\tilde \mY_i}} \cdot \mI_i (\mI - \mM_t) \cdot \nabla_\mZ \Ls + \frac{\partial \Ls(\mI_o \mM_t \tilde \mY, \mI_o \mM_t \mY)}{\partial \tilde \mY}$,
        \UNTIL{$\nabla_\mZ \Ls$ converges or iterations exhaust}
        \STATE $\vtheta = \vtheta - \eta \cdot \frac{\partial{\tilde \mY}}{\partial{\vtheta}} \cdot \nabla_\mZ \Ls$
    \end{algorithmic}
\end{algorithm}

Algorithm~\ref{algo:lignn} describes a method of Label-inputted Implicit GNN (LI-GNN) which combines naive forward iteration, implicit differentiation, and masked label strategy.
LI-GNN concatenates pseudo labels with the feature matrix $\mX$ and feeds them with the adjacency matrix $\mA$ into its backbone GNN $f_\vtheta$.
From lines 2 to 9, LI-GNN iteratively feeds pseudo labels into the backbone until it obtains the equilibrium $\hat \mY$ which is also used for inference after training.
The inputted pseudo labels are not necessarily masked here because the forward pass and the backward pass of IGNN are decoupled as long as the equilibrium is obtained.
In lines 10, MLS is applied to prevent label leakage and guarantee well-posedness.
In line 11, LI-GNN inputs its masked equilibrium state to cast an additional forward iteration with the gradients taped.
From line 12 to 15, LI-GNN iterate Equation~\ref{eq:id} to obtain the equilibrium gradients $\nabla_\mZ \Ls$.
During looping, gradients $\mI_i \mM_t \nabla_\mZ \Ls$ are omitted because of ground-truth labels overwriting in line 6.
In line 12, network weights $\vtheta$ get updated according to their gradients back-propagated.

\section{Experiments}\label{sec:experiments}

This section contains an analysis of the well-posedness of LI-GNN with experiments on five real-world datasets and a synthetic dataset.
Then, node classification experiments are conducted on another synthetic dataset and six real-world datasets to verify the superiority of LI-GNN\@.~\footnote{%
	Code for all experiments is released in~\url{Supplementary Material}.
}
Details of datasets and hyperparameters are in \Appref{app:exp}.

\subsection{Well-posedness analysis}

\begin{figure*}[t]
\centering
\includegraphics[width=0.8\linewidth]{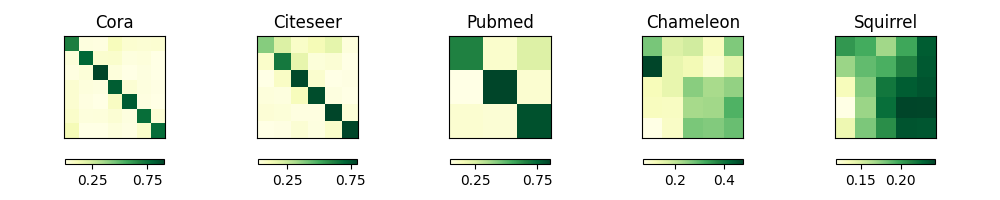}
\caption{Label transitional matrices of the heterophilic and homophilic graphs}\label{fig:datasets}
\end{figure*}

This experiment verifies how the ratio of masked labels influences LI-GCN's well-posedness using five real-world datasets (with their public splits in parentheses): Cora (5.17\%/18.46\%/36.93\%), Citeseer (3.61\%/15.03\%/30.06\%), Pubmed (0.30\%/2.54\%/5.07\%)~\cite{DBLP:journals/aim/SenNBGGE08}, Chameleon (47.96\%/32.02\%/20.03\%), and Squirrel (47.99\%/31.99\%/20.02\%)~\cite{DBLP:journals/compnet/RozemberczkiAS21}.
The label transition matrices of these five graphs are illustrated in Figure~\ref{fig:datasets}, describing the probability nodes with one label connecting to nodes with another.
As can be seen from the figure, most adjacent nodes in Cora, Citeseer, and Pubmed are same-labelled.
Graphs where adjacent nodes share similar labels are termed homophilic.
Other graphs, such as Chameleon and Squirrel, are heterophilic, where distant information is required to classify nodes correctly.

Lipschitz constant $K$ is a metric of how well-posed a mapping $f: \vz \rightarrow \vz$ is, which is defined as $ K = \sup\limits_{\forall \vz_1, \vz_2} {\| f(\vz_1) - f(\vz_2) \|} / {\| \vz_1 - \vz_2 \|}$.
By Banach's fixed-point theorem~\cite{kinderlehrer2000introduction}, if $K < 1$, the mapping $f$ is a contractor.
The smaller $K$ is, the faster iterating $f$ converges and the more well-posed the corresponding IGNN model is.
In the forward pass of an IGNN method, the measured mapping is $f_\vtheta$ in terms of $\mZ$.
In the backward pass, the measured mapping is $\frac{\partial f_\vtheta(\vz)}{\partial \vz}$ in terms of $\nabla_\mZ \Ls$.
Empirically, this work defines the stochastic Lipschitz constant $\hat K$ as
\begin{equation}\label{eq:lipconst}
    \hat K = \max\limits_{i = 1, \ldots, T} \frac{\| f(\vz_{i}) - \vz_{i} \|}{\| \vz_{i} - \vz_{i - 1} \|}
\end{equation}
for a $(T+1)$-length sequence $z_i, i = 0, \ldots, T$, to approximate $K$.

\begin{figure*}[t]
\centering
\includegraphics[width=1.0\linewidth]{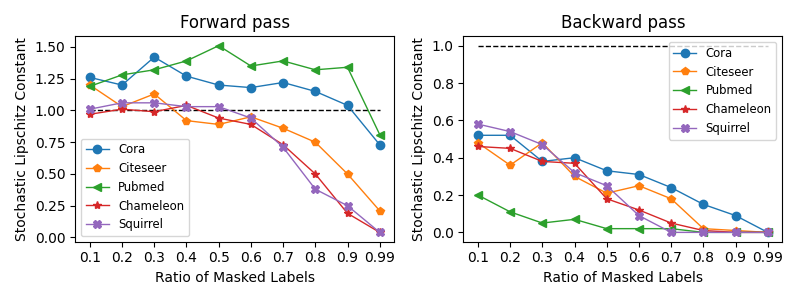}
\caption{Averaged stochastic Lipschitz constants in 10 runs of node classification}\label{fig:wellposed}
\end{figure*}

Figure~\ref{fig:wellposed} depicts mean stochastic Lipschitz constants in 10 runs of LI-GNN on the five datasets with different ratios $\beta$ of masked labels.
Each constant in a run is computed via Equation~\ref{eq:lipconst} with the sequence of $\mZ$ or $\nabla_\mZ \Ls$ retrieved from the model exceeding the highest validation f1-micro score.
Although $\hat K = K$ hardly holds with limited samples, there are clear downward trends of $\hat K$ as the ratio $\beta$ of masked labels rises, both in the forward pass and the backward pass.
In conclusion, a large mask rate $\beta$ suppresses the stochastic Lipschitz constant $\hat K$ of the backbone in IGNN, thus guaranteeing well-posedness.

\subsection{Label transition and label leakage}

This experiment demonstrates the effectiveness of the masked label strategy in preventing label leakage.
The synthesized graph is the Weekday~\cite{Luo2022InferringFR}, where each node represents a day.
A node is featured with an 8-dimensional vector encoding its date.
The label is its day of the week.
The graph is constructed by connecting the nearest nodes measured with Euclidean distances of features.
As the original article~\cite{Luo2022InferringFR} suggests, it is hard to judge a node's label in the Weekday graph purely by its features.
Classifiers have to capture the heterophilic label transitioning patterns among adjacent nodes.
The dataset is split into 60\%, 20\%, and 20\% for training, validation, and testing, respectively, as \citet{DBLP:conf/iclr/PeiWCLY20} suggests for full-supervised learning.

\begin{figure}[t]
\centering
\includegraphics[width=0.6\linewidth]{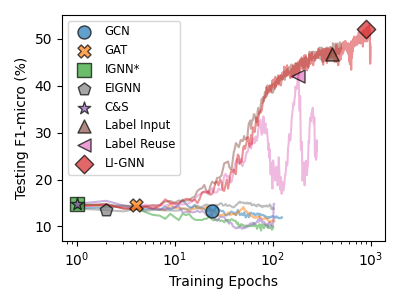}
\caption{A single run of node classification on the Weekday dataset}\label{fig:weekday}
\end{figure}

Figure~\ref{fig:weekday} illustrates curves of testing f1-micro scores for different methods on the Weekday in terms of training epochs, with testing f1-micro scores of the model exceeding the highest validation f1-micro scores marked.
As the figure depicts, performances of GCN, Graph Attention Network (GAT)~\cite{DBLP:conf/iclr/VelickovicCCRLB18}, IGNN*, and EIGNN are no better than a random guess because end-to-end methods cannot fully utilize label information.
C\&S also fails even though it is an LGNN method because it incorrectly assumes the homophilic label pattern by adopting nonparametric LPA\@.
Lable Input, Label Reuse, and LI-GNN capture the label transition patterns to some extent.
However, Label Reuse encounters rapid degradations when its score surpasses 30\% because of label leakage.
Although this method applies MLS in its forward pass, it leaks label information from iterating its trained network.
LI-GNN applies MLS after its forward iteration, rectifies the label leakage issue, and gains steady performance improvement as the training advances.

\subsection{Long-range dependencies}

This experiment evaluates the long-range dependencies capturing ability of models using the synthesized Chains datasets~\cite{DBLP:conf/nips/GuC0SG20} with 10 classes and 20 chains of nodes.
Class information of nodes on a chain is only provided sparsely in the feature of the end node.
The evaluated model is thus required to capture long-range dependencies to classify nodes correctly.
Different from \citet{DBLP:conf/nips/LiuKHWX21}, where each dataset is split into 5\%, 10\%, and 85\% for training, validation, and testing, this experiment samples 200 nodes for training.
It makes the label information sparser and the task more difficult when the chains are long.
The baselines are MLP, SGCs with 2, 10, and 100 layers, C\&S with 1, 10, and 100 iterations, Label Reuse with 0, 10, and 100 propagations, IGNN*, and EIGNN\@.

\begin{figure*}[t]
\centering
\includegraphics[width=1.0\linewidth]{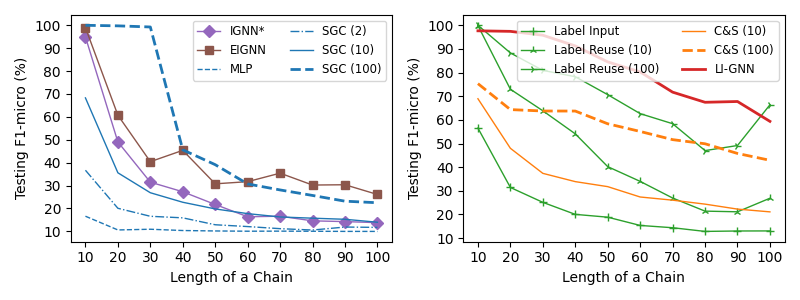}
\caption{Averaged testing F1-micro (\%) scores in 10 runs of 10-class node classification on the Chains datasets}\label{fig:chain}
\end{figure*}

As the results illustrated in Figure~\ref{fig:chain}, shallow GNN methods SGC (2) and SGC (10) cannot capture dependencies as chains lengthen.
If SGC is deep enough to have 100 layers, it can classify nodes with 100\% accuracy scores when the chain length is not larger than 30 because features of the end node on a chain can effectively propagate.
However, when the chain length surpasses 40, the performance of SGC (100) decreases rapidly because features after long-range propagations become over-smooth and undistinguishable~\cite{DBLP:conf/nips/GuC0SG20, DBLP:conf/nips/LiuKHWX21}.
Besides, IGNN* and EIGNN are stronger than SGC (10) in capturing long-range dependencies because of their infinite iterations, but they cannot surpass SGC (100) because infinite iterations do not sufficiently mean infinite effective range, as revealed in \citet{DBLP:journals/corr/abs-2210-08353}.
The over-smoothing issue in GNN and the limited-effective-range issue in IGNN reveal that label information is difficult to supervise end-to-end learning as the propagation range enlarges.
Therefore, explicitly utilizing label information like LGNN can alleviate this issue.
The performances of C\&S and Label Reuse steadily boost with increasing iterations that C\&S (100) and Label Reuse (100) surpass all end-to-end methods.
Moreover, with proper optimization, namely implicit differentiation, LI-GNN achieves the best performance.
In conclusion, LI-GNN combines the propagating advantage of LGNN and the learning advantage of IGNN, and thus it has the best long-range dependencies capturing ability.

\subsection{Real-world node classification}

\begin{table}\small
\caption{Mean ± stderr of testing F1-micro (\%) in 10 runs on real-world datasets}\label{tbl:acc}
\centering
	\begin{tabular}{lcccccc}
		\toprule
                                     &                    &                   &                 & \textbf{Amazon} & \textbf{Coauhthor} & \textbf{OGB-Arxiv} \\
                                     & \textbf{Chameleon} & \textbf{Squirrel} & \textbf{Flickr} & \textbf{Photo}  & \textbf{Physics}   & {\scriptsize (transductive)} \\
		\textbf{\#Nodes}             & 2277               & 5201              & 89250           & 7650            & 34493              & 169343  \\
		\textbf{\#Edges}             & 36051              & 216933            & 449878          & 119081          & 247962             & 1166243 \\
		\textbf{Average Degree}      & 31.67              & 83.42             & 10.08           & 31.13           & 14.38              & 13.77   \\
		\textbf{\#Features}          & 2325               & 2089              & 500             & 745             & 8415               & 128     \\
		\textbf{\#Classes}           & 5                  & 5                 & 7               & 8               & 5                  & 40      \\
		\textbf{Intra-class Edges}   & 23.39\%            & 22.34\%           & 31.95\%         & 82.72\%         & 93.14\%            & 65.51\% \\
		\midrule
        GCN        & 39.28±3.65          & 26.87±2.19          & 38.36±6.28          & 86.90±2.75          & 91.10±1.97           & 70.70±0.18 \\
		GCN + C\&S & 41.04±3.80          & 27.10±2.11          & 38.74±6.20          & 87.21±2.87          & 91.75±1.76           & 71.50±0.12 \\
        L-GCN      & \textbf{42.46±3.90} & 28.35±2.49          & 43.01±4.79          & 88.54±2.62          & \textbf{91.84±1.34}  & 71.97±0.21 \\
        LI-GCN     & 42.18±2.72          & \textbf{28.95±2.23} & \textbf{43.47±3.12} & \textbf{88.58±2.95} & 91.83±1.21           & \textbf{71.99±0.30} \\
		\midrule
        IGNN*       & 35.40±2.36          & 24.97±1.28          & 28.42±5.54          & 87.19±1.95          & 90.92±1.71          & 70.13±0.30 \\
		SotA        & 39.24±2.02          & 28.86±2.08          & 39.65±4.25          & 87.98±2.23          & 92.69±1.35          & 71.92±0.20 \\
		SotA + C\&S & 40.04±2.37          & 28.77±2.06          & \textbf{42.21±2.17} & 88.00±1.97          & 92.55±1.47          & 72.25±0.18 \\
        L-SotA      & 41.41±3.30          & 29.03±2.03          & 42.08±4.60          & 88.40±3.01          & 92.65±1.14          & \textbf{72.86±0.14} \\
        LI-SotA     & \textbf{41.53±3.58} & \textbf{29.53±1.58} & 40.51±6.03          & \textbf{88.54±2.49} & \textbf{92.81±1.13} & 72.54±0.23 \\
		\bottomrule
	\end{tabular}
\end{table}

This experiment evaluates the real-world node classification performance of methods on six datasets, including Chameleon, Squirrel~\cite{DBLP:journals/compnet/RozemberczkiAS21}, Flickr~\cite{DBLP:conf/iclr/ZengZSKP20}, Amazon Photo~\cite{DBLP:conf/sigir/McAuleyTSH15}, Coauthor Physics~\cite{DBLP:journals/qss/WangSHWDK20}, and OGB-Arxiv~\cite{DBLP:conf/nips/HuFZDRLCL20}.
These datasets are different in many aspects, including scale measured by numbers of nodes and edges, density measured by node degrees, and homophily measured by intra-class edge rate~\cite{DBLP:journals/corr/abs-2002-06755}, as summarized in the upper part of Table~\ref{tbl:acc}.
Datasets except OGB-Arxiv are split following the semi-supervised and inductive settings~\cite{DBLP:conf/iclr/ChenMX18} to make label information sparsely distributed.
In detail, only 10 nodes from each class are sampled for training.
500 and 1000 random nodes form the validation and testing set.
Edges associated with the latter 1500 nodes are masked during training.
Apart from IGNN*, evaluated models include GNN, GNN with C\&S, GNN with Label Input or Label Reuse (L-GNN), and LI-GNN\@.
The backbone GNN is either GCN or a state-of-the-art GNN (SotA) chosen from GAT, JKNet~\cite{DBLP:conf/icml/XuLTSKJ18}, and GCNII~\cite{DBLP:conf/icml/ChenWHDL20}.
EIGNN is not considered in this experiment because its preprocessing is too time-consuming when the number of nodes reaches 10000.

Every model runs 10 times on a dataset to collect the testing f1-micro score with the highest validation score in each run.
The averaged testing f1-micro score is reported in Table~\ref{tbl:acc}.
Compared with IGNN, LGNN methods are superior because they can be widely applied to varying GNN backbones, achieving competitive performances in various real-world tasks.
Among LGNN methods, LI-GNN behaves best in most cases because its label propagation is both adaptive and distant.

\section{Conclusions}

This article unifies LGNN and IGNN, proposing a combined method LI-GNN with long and adaptive label propagation and arbitrary backbone network.
With excessive experiments, its effectiveness is verified and its properties are revealed.
Its superiority further examines the inspiring effect of the connection between LGNN and IGNN.

\begin{ack}
Use unnumbered first level headings for the acknowledgments. All acknowledgments
go at the end of the paper before the list of references. Moreover, you are required to declare
funding (financial activities supporting the submitted work) and competing interests (related financial activities outside the submitted work).
More information about this disclosure can be found at: \url{https://neurips.cc/Conferences/2023/PaperInformation/FundingDisclosure}.

Do {\bf not} include this section in the anonymized submission, only in the final paper. You can use the \texttt{ack} environment provided in the style file to autmoatically hide this section in the anonymized submission.
\end{ack}

\appendix
\section{Derivation of Table 1: Rewriting LGNN methods in the form of IGNN}
\label{app:lgnns}
    
This section prooves that LPA~\cite{DBLP:conf/icml/ZhuGL03}, ResLPA~\cite{DBLP:conf/iccai/LuoHCZ21}, 3ference~\cite{Luo2022InferringFR}, Correct of C\&S~\cite{DBLP:conf/iclr/HuangHSLB21}, and Label Input~\cite{DBLP:conf/ijcai/ShiHFZWS21} for GNN with linear propagation layers are IGNNs with pseudo labels $\tilde \mY_t$ to be the equilibrium, where pseudo labels $\tilde \mY_t$ are defined as predictions $\hat \mY$ overwritten by known ground-truth labels $\mM_t \mY$, as
\begin{equation}\label{eq:y}
    \tilde \mY_t \triangleq \mM_t \mY + \mM_{\bar t} \hat \mY.
\end{equation}

\begin{remark}[LPA is an IGNN*]\begin{proof}
    Converged prediction $\hat \mY$ of LPA admits
    \[
        \hat \mY = \tilde \mA_\lambda \tilde \mY_t.
    \]
    Plugging it into (\ref{eq:y}) gets
    \[
        \tilde \mY_t = \mM_{\bar t} \tilde \mA_\lambda \tilde \mY_t + \mM_t \mY,
    \]
    where $\mM_{\bar t} \tilde \mA_\lambda$ can be regarded as a new normalized adjacency matrix, and $\mM_t \mY$ is the bias term omitted in the definition of IGNN*.
    Thus, LPA is an IGNN* without node features.
\end{proof}\end{remark}

\begin{remark}[ResLPA is an IGNN* $\circ$ GNN]\begin{proof}
    ResLPA leverages node features to fix propagated labels.
    With a GNN as its residual estimator, ResLPA is to iterate
    \[
        \hat \mY = (1 - \lambda) \tilde \mY_t + \lambda \cdot (\tilde \mA \tilde \mY_t + f_\vtheta(\mA, \mX))
        = \tilde \mA_\lambda \tilde \mY_t + \lambda f_\vtheta(\mA, \mX).
    \]
    Plugging it into (\ref{eq:y}) gets
    \[
        \tilde \mY_t = \mM_{\bar t} \tilde \mA_\lambda \tilde \mY_t + \lambda \mM_{\bar t} f_\vtheta(\mA, \mX) + \mM_t \mY,
    \]
    an IGNN* stacked upon a GNN that outputs $\lambda \mM_{\bar t} f_\vtheta(\mA, \mX)$.
\end{proof}\end{remark}

\begin{remark}[3ference is an IGNN with a $\mbox{GCN}^{(1)}$ backbone]\begin{proof}
    3ference adaptively learns label transition patterns with a GNN to avoid heuristic assumptions.
    It combines node features and known labels in the neighbourhood to predict the target labels.
    Assuming that the predictor is a GNN $f_\vtheta$, which is suggested to gather information from one-hop neighbourhoods without label information from the central node, 3ference is to iterate
    \[
        \hat \mY = f_\vtheta(\mA - \mbox{diag}(\mA), \begin{bmatrix} \tilde \mY_t & \mX \end{bmatrix}).
    \]
    Plugging it into (\ref{eq:y}) gets
    \[
        \tilde \mY_t = \mM_{\bar t} f_\vtheta(\mA - \mbox{diag}(\mA), \begin{bmatrix} \tilde \mY_t & \mX \end{bmatrix}) + \mM_t \mY.
    \]
    $\mM_{\bar t}$ is a diagonal 0-1 matrix that can pass through element-wise activators such as LeakyReLU and Sigmoid to merge into GNN\@.
    Thus, 3ference is an IGNN with a $\mbox{GCN}^{(1)}$ backbone.
\end{proof}\end{remark}

\begin{remark}[Correct of C\&S is an IGNN*]\begin{proof}
    The Correct phase of C\&S propagates residuals between known labels $\mM_t \mY$ and the base prediction $\hat \mY_0 = \mbox{MLP}(\mX)$, as
    \[\begin{aligned}
        \hat \mY &= \tilde \mA_{\lambda_c} (\mM_{\bar t} (\hat \mY - \hat \mY_0) + \mM_t (\mY - \hat \mY_0)) + \hat \mY_0 \\
        &= \tilde \mA_{\lambda_c} (\tilde \mY_t - \hat \mY_0) + \hat \mY_0 \\
        &= \tilde \mA_{\lambda_c} \tilde \mY_t + (\mI - \tilde \mA_{\lambda_c}) \hat \mY_0 \\
    \end{aligned}\]
    Plugging it into (\ref{eq:y}) gets
    \[
        \tilde \mY_t = \mM_{\bar t} \tilde \mA_{\lambda_c} \tilde \mY_t + \mM_{\bar t} (\mI - \tilde \mA_{\lambda_c}) \hat \mY_0 + \mM_t \mY,
    \]
    Regarding $\mM_{\bar t}(\mI - \tilde \mA_{\lambda_c}) \hat \mY_0$ as the diffusion of the base prediction $\hat \mY_0$, Correct is an IGNN* stacked upon it.
\end{proof}\end{remark}

\begin{remark}[Label Input (SGC) is an IGNN]\begin{proof}
    Label Input randomly splits the known labels $\mM_t \mY$ into two exclusive parts in every epoch, inputting one part $\mM_i \mY$ into GNNs with another $\mM_o \mY$ to supervise the output.
    SGC with Label Input can be formulated as
    \[
        \hat \mY = (\mM_{\bar t} + \mM_o) f_\vtheta(\mA, \begin{bmatrix} \mM_i \mY & \mX \end{bmatrix}) + \mM_i \mY.
    \]
    \cite{DBLP:journals/corr/abs-2110-07190} proves that with linear propagation layers, the deterministic objective is to minimize the difference between known labels $\mM_t \mY$ and
    \[
        \sigma([\tilde \mA - \mbox{diag}(\tilde \mA)] \mM_t \mY \mW_y + \tilde \mA \mX \mW_x),
    \]
    with a regularization term omitted.
    Thus, Label Input for an SGC can be converted to an IGNN, as
    \[\begin{aligned}
        \tilde \mY_t &= \mM_{\bar t} \sigma([\tilde \mA - \mbox{diag}(\tilde \mA)] \mM_t \tilde \mY_t \mW_y + \tilde \mA \mX \mW_x) + \mM_t \mY \\
        &= \sigma(\mM_{\bar t} [\tilde \mA - \mbox{diag}(\tilde \mA)] \mM_t \tilde \mY_t \mW_y + \mM_{\bar t} \tilde \mA \mX \mW_x) + \mM_t \mY, \\
    \end{aligned}\]
    with $\sigma$ is element-wise such as LeakyReLU.
\end{proof}\end{remark}

\section{Proof of Theorem 1: Sufficient condition for well-posedness of IGNN with a $\mbox{GCN}^{(1)}$ backbone}
\label{app:wellposed}
\begin{proof}

IGNN with a one-hop convolutional backend is denoted as
\[
    f(\mZ) = \sigma(\tilde \mA \mZ \mW_z + \mP \mX \mW_x),
\]
where $\sigma$ is the Softmax or Sigmoid function for node classification, $\tilde \mA$ is an $n \times n$ normalized adjacency matrix, $\mP$ is an $n \times n$ matrix that is not necessarily the same as $\tilde \mA$.
When $\mP = \tilde \mA$, $f$ is a $\mbox{GCN}^{(1)}$.
When $\mP = \mI$, $f$ is an IGNN*.

For any $\mZ_1$ and $\mZ_2$, with $\| \mW_z \|_F < 1 / n$,
\begin{align}
    &\| f(\mZ_1) - f(\mZ_2) \|_F \nonumber \\
    &= \| \sigma(\tilde \mA \mZ_1 \mW_z + \mP \mX \mW_x) - \sigma(\tilde \mA \mZ_2 \mW_z + \mP \mX \mW_x) \|_F \nonumber \\
    &\le \| (\tilde \mA \mZ_1 \mW_z + \mP \mX \mW_x) - (\tilde \mA \mZ_2 \mW_z + \mP \mX \mW_x) \|_F \label{eq:softmax} \\
    &= \| \tilde \mA (\mZ_1 - \mZ_2) \mW_z \|_F \nonumber \\
    &\le \| \tilde \mA \|_2 \cdot \| (\mZ_1 - \mZ_2) \mW_z \|_F \label{eq:norm} \\
    &\le n \| \tilde \mA \|_{\mbox{max}} \| \cdot \| \mZ_1 - \mZ_2 \|_F \cdot \| \mW_z \|_F \nonumber \\
    &\le n \| \mZ_1 - \mZ_2 \|_F \| \mW_z \|_F \nonumber \\
    &\le \| \mZ_1 - \mZ_2 \|_F. \nonumber \\ \nonumber
\end{align}
\noindent
(\ref{eq:softmax}) holds because Softmax and Sigmoid is nonexpansive ~\cite{DBLP:journals/corr/abs-1704-00805}.
(\ref{eq:norm}) holds because of the Lemma B.3 in \cite{DBLP:conf/iclr/ZouLG20}.
Thus, $f$ is a contractor.

By Banach's fixed-point theorem~\cite{kinderlehrer2000introduction}, iterating $f$ from an arbitrary start point is guaranteed to converge to the existing and unique fixed point.
Thus, IGNN with $f$ as the backbone is well-posed.
\end{proof}

\section{Experiment details}
\label{app:exp}

The real-world datasets in this work are retrieved by PyG~\cite{Fey/Lenssen/2019} and DGL~\cite{wang2019dgl}.
In detail, Chameleon and Squirrel~\cite{DBLP:journals/compnet/RozemberczkiAS21} are heterophilic networks of web pages on Wikipedia.
They are classified by monthly traffic and are connected if mutual links exist.
Flickr~\cite{DBLP:conf/iclr/ZengZSKP20} is a heterophilic dataset of categorized online images connected considering their common properties.
Cora, Citeseer, Pubmed~\cite{DBLP:journals/aim/SenNBGGE08}, Amazon Photo, Coauthor Physics~\cite{DBLP:journals/corr/abs-1811-05868}, and OGB-Arxiv~\cite{DBLP:conf/nips/HuFZDRLCL20} are homophilic networks.
Cora, Citeseer, Pubmed, and OGB-Arxiv are networks of typed articles with citations represented as graph edges.
Amazon Photo is a segment of the Amazon co-purchase graph~\cite{DBLP:conf/sigir/McAuleyTSH15}, constructed with on-sale goods and their co-purchase relationship.
Coauthor Physics is a graph with authors labelled by study fields and connected by co-authorship, based on the Microsoft Academic Graph (MAG)~\cite{DBLP:journals/qss/WangSHWDK20}.

\begin{table}\small
\caption{Hyperparameters for grid searching}\label{tbl:tune}
\centering
	\begin{tabular}{lcc}
		\toprule
Method & Hyperparameters & Searching Scope \\
		\midrule
GAT & \#Attention Heads & $\{ 2, 4, 8 \}$ \\
JKNet & \#Layers & $\{ 2, 4, 8 \}$ \\
GCNII & (\#Layers, $\alpha$, $\theta$) & $\{2, 4, 8\} \times \{0.1, 0.3, 0.5\} \times \{0.5, 1.0, 1.5\}$ \\
Label Input \& Label Reuse & ($\alpha$, \#Iterations) & $\{0.25, 0.50, 0.75\} \times \{0, 1, 3\} $ \\
C\&S & ($\lambda_c$, $\lambda_s$) & $\{0.1, 0.2, 0.3\} \times \{0.1, 0.2, 0.3\}$ \\
LI-GNN & (forward mask flag, $\beta$) & $\{\mbox{true}, \mbox{false}\} \times \{0.25, 0.50, 0.75\}$ \\
		\bottomrule
	\end{tabular}
\end{table}

\begin{table}\small
\caption{Searched SotA and hyperparameters}\label{tbl:sota}
\centering
	\begin{tabular}{lcccc}
		\toprule
        Dataset & Backbone & C\&S & Label Tricks & LI-GNN \\
		\midrule
        Chameleon        & GCN                & (0.2, 0.1) & (0.50, 1) & (false, 0.25) \\
                         & JKNet(4)           & (0.1, 0.2) & (0.75, 3) & (true, 0.75) \\
        Squirrel         & GCN                & (0.2, 0.2) & (0.75, 1) & (true, 0.75) \\
                         & JKNet(8)           & (0.2, 0.1) & (0.25, 3) & (true, 0.50) \\
        Flickr           & GCN                & (0.3, 0.3) & (0.75, 3) & (false, 0.25) \\
                         & GCNII(2, 0.1, 0.5) & (0.3, 0.3) & (0.75, 0) & (false, 0.25) \\
        Amazon Photo     & GCN                & (0.1, 0.1) & (0.25, 0) & (false, 0.25) \\
                         & GCNII(4, 0.1, 0.5) & (0.3, 0.3) & (0.50, 1) & (true, 0.50) \\
        Coauthor Physics & GCN                & (0.1, 0.3) & (0.25, 0) & (true, 0.25) \\
                         & GCNII(8, 0.3, 1.5) & (0.1, 0.1) & (0.25, 0) & (true, 0.50) \\
        OGB-Arxiv        & GCN                & (0.2, 0.2) & (0.75, 0) & (true, 0.50) \\
                         & GCNII(4, 0.1, 0.5) & (0.2, 0.2) & (0.75, 0) & (true, 0.50) \\
		\bottomrule
	\end{tabular}
\end{table}

The code for IGNN* and EIGNN is retrieved from the official code repository of EIGNN.~\footnote{\url{https://github.com/liu-jc/EIGNN}}
Other baselines are imported from PyG or implemented using PyTorch~\cite{NEURIPS2019_9015}.
The dimension of hidden representations for all methods is 64, except for EIGNN where the dimension equals that of input and for OGB-Arxiv where the dimension is 256.
All MLPs are 3-layered.
All GCNs and GATs are 2-layered.
Other hyperparameters are grid searched, as summerized in Table~\ref{tbl:tune}.
The state-of-the-art method for each dataset and the searched hyperparameters for C\&S, Label Tricks (Label Input and Label Reuse), and LI-GNN are in Table~\ref{tbl:sota}.

If not specially mentioned, the Adam optimizer~\cite{DBLP:journals/corr/KingmaB14} is adopted for training with the learning rate of 0.01.
The early stopping strategy is applied to terminate the training if a model fails to gain performance increase during 100 epochs.

{
    \small
    \bibliographystyle{apalike}
    \bibliography{gqn.bib}
}

\end{document}